\documentclass[conference]{IEEEtran}
\IEEEoverridecommandlockouts
\usepackage{hyperref}
\hypersetup{colorlinks,allcolors=black}
\usepackage{cite}
\usepackage{amsmath,amssymb,amsfonts}
\usepackage{algorithmic}
\usepackage{graphicx}
\usepackage{textcomp}
\usepackage{xcolor}
\usepackage{float}
\usepackage{diagbox}
\usepackage{soul}
  
\def\BibTeX{{\rm B\kern-.05em{\sc i\kern-.025em b}\kern-.08em
    T\kern-.1667em\lower.7ex\hbox{E}\kern-.125emX}}
\begin{document}

\title{Mechatronic Investigation of Wound Healing Process by Using Micro Robot
}

\makeatletter
\newcommand{\linebreakand}{%
  \end{@IEEEauthorhalign}
  \hfill\mbox{}\par
  \mbox{}\hfill\begin{@IEEEauthorhalign}
}
\makeatother

\author{\IEEEauthorblockN{Abdurrahim Yilmaz}
\IEEEauthorblockA{\textit{Yildiz Technical University} \\
\textit{Mechatronics Engineering}\\
Istanbul, Turkey \\
a.rahim.yilmaz@gmail.com}
\and
\IEEEauthorblockN{Ali Anil Demircali}
\IEEEauthorblockA{\textit{Imperial College London} \\
\textit{The Hamlyn Centre}\\
London, The United Kingdom \\
a.demircali@imperial.ac.uk}
\and
\IEEEauthorblockN{Serra Ozkasap}
\IEEEauthorblockA{\textit{Dokuz Eylul University} \\
\textit{Industrial Engineering}\\
Izmir, Turkey \\
serraozkasap@gmail.com}
\linebreakand
\IEEEauthorblockN{Leyla Yorgancioglu}
\IEEEauthorblockA{\textit{Enka Schools} \\
\textit{}\\
Istanbul, Turkey \\
leyla.yorgancioglu@stu.enka.k12.tr}
\and
\IEEEauthorblockN{Huseyin Uvet}
\IEEEauthorblockA{\textit{Yildiz Tehnical University} \\
\textit{Mechatronics Engineering}\\
Istanbul, Turkey \\
huvet@yildiz.edu.tr}
\and
\IEEEauthorblockN{Gizem Aydemir}
\IEEEauthorblockA{\textit{Sorbonne Université} \\
\textit{Institut des Systèmes Intelligents et de Robotique}\\
Paris, France \\
aydemir@isir.upmc.fr}
}

\maketitle

\begin{abstract}
The purpose of this study is to find ideal forces for reducing cell stress in wound healing process by micro robots. Because of this aim, we made two simulations on COMSOL® Multiphysics with micro robot to find correct force. As a result of these simulation, we created force curves to obtain the minimum force and friction force that could lift the cells from the surface will be determined. As the potential of the system for two micro robots that have 2 mm x 0.25 mm x 0.4 mm dimension SU-8 body with 3 NdFeB that have 0.25 thickness and diameter, simulation results at maximum force in the x-axis calculated with 4.640 mN, the distance between the two robots is 150 $\mu$m.
\end{abstract} 


\section{Introduction}

Micro robots are tools with minimal human impact that perform operations in many different areas. Since they are not connected to any system, they have the potential to revolutionize the field of health due to their highly maneuverable structures and functioning such as minimizing invasive procedures in disease diagnosis \cite{martel2013magnetic}. Studies and rapidly advancing micro robot technology show that micro robots have the potential to perform tasks. These tasks are currently difficult or impossible such as Sitti has discussed about microrobotic devices that can navigate inside human body \cite{sitti2009voyage}. It is also predicted that it will lead to the development of treatments that have not yet been researched in the future.

Micro robotic research on health sector initially concentrated on the principles of microscopic locomotion methods and their dynamic interactions with their surrounding fluids and surfaces \cite{sitti2017mobile}. Within the scope of studies in the field of micro robots, different movement methods have emerged due to the innovations developed in the design, production and control systems of micro robots. One of the areas where micro robots are used efficiently is cell studies such as cell migration. Various in vitro methods have been used to analyze cell migration, because cell migration plays an essential role in wound healing processes \cite{cappiello2018novel}. To improve understanding of wound healing processes, chemical, physical or biological parameters should be tested for their ability to induce or inhibit the endothelial cell migration.

Cell migration is affected by many intracellular pathways and extracellular stimuli such as growth factors, mechanical stress, and chemical interactions \cite{alberts2002extracellular}. It has been demonstrated in many studies that dynamic cell cultures present in vivo environment in more detail. As a result, dynamic cell cultures, especially those integrated into microfluidic chips, exhibit more natural properties than cultures within microfluidic chips.
Micro robots that used in healthcare need to perform their tasks in fluid environments with low Reynolds number \cite{purcell1977life}. Because of their sizes, they tend to face high friction forces and inertia when operating in these environments. It is possible to transmit power to the micro robot via wireless network, instead of the micro robot working with little or no power storage \cite{nelson2010microrobots}. While performing this process, magnetic fields are used in this study.

We have two issues in the present study case. Firstly, the cellular studies want standardization for micro robotic wounding dimension. The methods that used for scratching cells have different variations and these methods could give some different micro robotic wounding dimensions. In the other hand, these problem can solved many methods like laser applications. However, applications of these methods give another problem that stressed cells. The stress effect can be minimized by creating a uniform wound of equal diameter and correct forces for the experimental and control groups. Thus, the effect of the medicines on the rate of recovery will be interpreted more accurately. 

The aim of this study is to present the force curves created by the base micro robot on the wounding micro robot. The curves are calculated according to the distance between these robots. The magnitude of the force required to micro robotic wounding will be obtained with a simulation-based force curve. Thus, the minimum force that can lift the cells from the surface will be determined. Moreover, theoretical background of the simulation was added the mathematical model. This technique for such applications lies in the standardized wound healing model capable of generating various wound shapes with uniform widths within microfluidic chip platform. With the method examined in this study, standardized wound models that can be used in different cell biology studies can be realized.

\section{Mathematical Model}
In this section, the mathematical model for the creating wound by using the micro robot subject to the liquid environment is presented. The micro robot contains 3 NdFeB (neodymium) cylindrical-shaped permanent magnets with a 0.25 mm diameter and 0.25 mm thickness and is manipulated diamagnetically. The major parameters of the system are shown in Table \ref{tbl:mp}.

\begin{table}[ht]
\small
  \caption{\ Major parameters of the system}
  \label{tbl:mp}
  \begin{tabular*}{0.48\textwidth}{@{\extracolsep{\fill}}lll}
    \hline
    Symbol & Parameter & Unit \\
    \hline
    $F_B$ & Buoyancy Force & N \\
    $F_{M}$ & Magnetic Force & N \\
    $F_G$ & Gravitational Force & N \\
    $F_{ES}$ & Electrostatic Force & N \\
    $F_A$ & Adhesive Force & N \\
    $F_D$ & Drag Force & N \\
    $N$ & Reaction Normal Force & N \\
    $Re$ & Reynolds Number & - \\
    $A$ & Cross-Sectional Area & $m^2$ \\
    $\mu_r$ & Magnetic Permeability of Space &  \\
    $\mu_0$ & Magnetic Conductivity Number & Wb/Am \\
    $H$ & Magnetic Field & A/m \\
    $M_{dia}$ & Magnetization Vector & A/m \\
    $\chi$ &Magnetic Susceptibility Factor & - \\
    $B$ & Magnetic Flux Density & T \\
    \hline
  \end{tabular*}
\end{table}

The continuity equation is written in Eq. \ref{eqn:ce} for incompressible flows along the channel:

\begin{equation}\label{eqn:ce}
    \nabla.u=0
\end{equation}

The Reynolds Number is expressed in Eq. \ref{eqn:rn}\cite{taitel1976model}. 

\begin{equation}\label{eqn:rn}
    R_{e}=\frac{\rho \cdot v \cdot D_{h}}{\mu}
\end{equation}

The momentum equation is described as \cite{sakiadis1961boundary}:

\begin{equation}\label{eqn:me}
\rho \frac{\partial \vec{v}}{\partial t}=-\vec{\nabla} P+\rho \vec{g}+\eta \nabla^{2} \vec{v}+\overrightarrow{F_{M}}
\end{equation}

In this equation, $\rho$ is density, $v$ is flow velocity, $\mu$ is viscosity $P$ is pressure. The force that is produced by magnet on the micro-robot is applied in the channel and affected force is given in Eq. \ref{eqn:df}.

\begin{equation}\label{eqn:df}
d \vec{F}=\mu_{0} \nabla(\vec{M} . \vec{H})
\end{equation}

The magnetization of the magnet is M, and the magnetic field strength is H in this formula. On the micro-robot, the total magnetic force is expressed in Eq. \ref{eqn:tf}.

\begin{equation}\label{eqn:tf}
\overrightarrow{F_{m}}=\int_{V} d \vec{F} d V
\end{equation}

The forces in cylindirical coordinates are written as respectively Eq. \ref{eqn:fz},\ref{eqn:fy},\ref{eqn:fx}:

\begin{equation}\label{eqn:fz}
\vec{F}_{z}=\int_{V} d \vec{F}_{z}-\mu_{0} \int_{V} \frac{\partial}{\partial z}\left(M_{0} H_{z}\right)(\mathrm{Z} \text { axis})
\end{equation}

\begin{equation}\label{eqn:fy}
\vec{F}_{y}=\int_{V} d \vec{F}_{y}-\mu_{0} \int_{V} \frac{\partial}{\partial y}\left(M_{0} H_{z}\right)(\mathrm{Y}  \text { axis})
\end{equation}

\begin{equation}\label{eqn:fx}
\vec{F}_{x}=\int_{V} d \vec{F}_{x}-\mu_{0} \int_{V} \frac{\partial}{\partial x}\left(M_{0} H_{z}\right)(\mathrm{X} \text { axis})
\end{equation}

The force calculation in cylindirical coordinates is made with the uniform medium assumption. Consequently, the diamagnetic force is defined in Eq. \ref{eqn:dff}.

\begin{equation}\label{eqn:dff}
d \textbf{F}=M_{d i a}(\nabla \textbf{B}) d V
\end{equation}

The diamagnetic force in x, y, z axis are as follows \cite{waldron1966diamagnetic}:

\begin{equation}\label{eqn:fdx}
F_{\text {dia},x}=\frac{X_{\text {dia}}}{2 \mu_{0}} \iiint_{V}\left(\frac{\partial|B|^{2}}{\partial x}\right) d V 
\end{equation}

\begin{equation}\label{eqn:fdy}
F_{\text {dia},y}=\frac{X_{\text {dia}}}{2 \mu_{0}} \iiint_{V}\left(\frac{\partial|B|^{2}}{\partial y}\right) d V
\end{equation}

\begin{equation}\label{eqn:fdz}
F_{\text {dia},z}=\frac{X_{\text {dia}}}{2 \mu_{0}} \iiint_{V}\left(\frac{\partial|B|^{2}}{\partial z}\right) d V
\end{equation}

According to Ostrogradsky’s divergence law, the diamagnetic forces can be simplified, and they describes as \cite{katz1979history}:

\begin{equation}
F_{\text{dia},x} =\frac{\chi_{d}}{\mu_{0}} \iint_{S}\|B\|^{2} n_{x} d s
\end{equation}

\begin{equation}
F_{\text{dia},y} =\frac{\chi_{d}}{\mu_{0}} \iint_{S}\|B\|^{2} n_{y} d s
\end{equation}

\begin{equation}
F_{\text{dia},z} =\frac{\chi_{d}}{\mu_{0}} \iint_{S}\|B\|^{2} n_{z} d s
\end{equation}

Buoyancy force, drag force, commonly known as non-gravitational forces, and gravitational force affect the micro-magnet of the micro-robot and are defined as:

\begin{equation}
F_{G}=m_{m} g
\end{equation}

\begin{equation}
F_{B}=V_{m} \rho_{f} g
\end{equation}

\begin{equation}
F_{D}=\frac{1}{2} c_{d} \rho_{f} A v^{2}
\end{equation}

$m_m$, $V_m$, $\rho_f$ and $C_d$ are the micro-magnet mass, volume, fluid density, and drag coefficient, respectively. The drag force coefficient function can be stated as \cite{payne1958calculations}:

\begin{equation}
C_{d}=\frac{24}{R e}\left(1+\frac{R e^{2 / 3}}{6}\right)
\end{equation}

Affected other forces on micro robot that are electro-static force, adhesive force, reaction normal force and friction force. The net force in x axis is sum of x component of magnetic force, friction force and drag force in Eq. \ref{eqn:fnetx}. The net force in z axis is sum of reaction normal force, z component of magnetic force, adhesive force, buoyancy force, electro-static force and gravitational force in Eq. \ref{eqn:fnetz}.

\begin{equation}\label{eqn:fnetz}
F_{n e t, z}=N+F_{M, z}+F_{A}-F_{G}-F_{E S}-F_{B} \\
\end{equation}

\begin{equation}\label{eqn:fnetx}
F_{n e t, x}=F_{M, x}-F_{f}-F_{D}
\end{equation}

\section{Simulation Results}
Two analyzes were carried out with the COMSOL® Multiphysics Version 5.3 (CPU License No:17076072), which is one of the finite element analysis programs, in order to determine the forces applied by the micro robots to each other during the wound opening process and to calculate the friction force. SU-8 is selected as the micro robot body. The characteristics of the two analyzes are shown in Table \ref{tbl:params}. 

\begin{table}[ht]
\small
  \caption{\ Electromagnetic and Friction Force Analysis Parameters}
  \label{tbl:params}
  \begin{tabular*}{0.48\textwidth}{@{\extracolsep{\fill}}lll}
    \hline
    Parameter & Electromagnetic Force & Friction Force \\
    \hline
    Study Type & Stationary & Time-Dependent \\
    Length & 500 $\mu$m & 1 second \\
    Step & 10 $\mu$m & 0.01 second \\
    Physiycs & mfnc & spf - solid - fsip \\
    \hline
  \end{tabular*}
\end{table}

\subsection{Electromagnetic Force Analysis}
In the first analysis, the analysis was performed by increasing the distance along the x-axis between the centerline of robots with 10 micrometers. The analysis was repeated with parametric sweep until the distance 500 micrometers with 10 micrometer steps. The analysis environment is shown in Fig. \ref{fgr:force_overview}.

\begin{figure}[ht]
\centering
  \includegraphics[height=5cm, width=8.8cm]{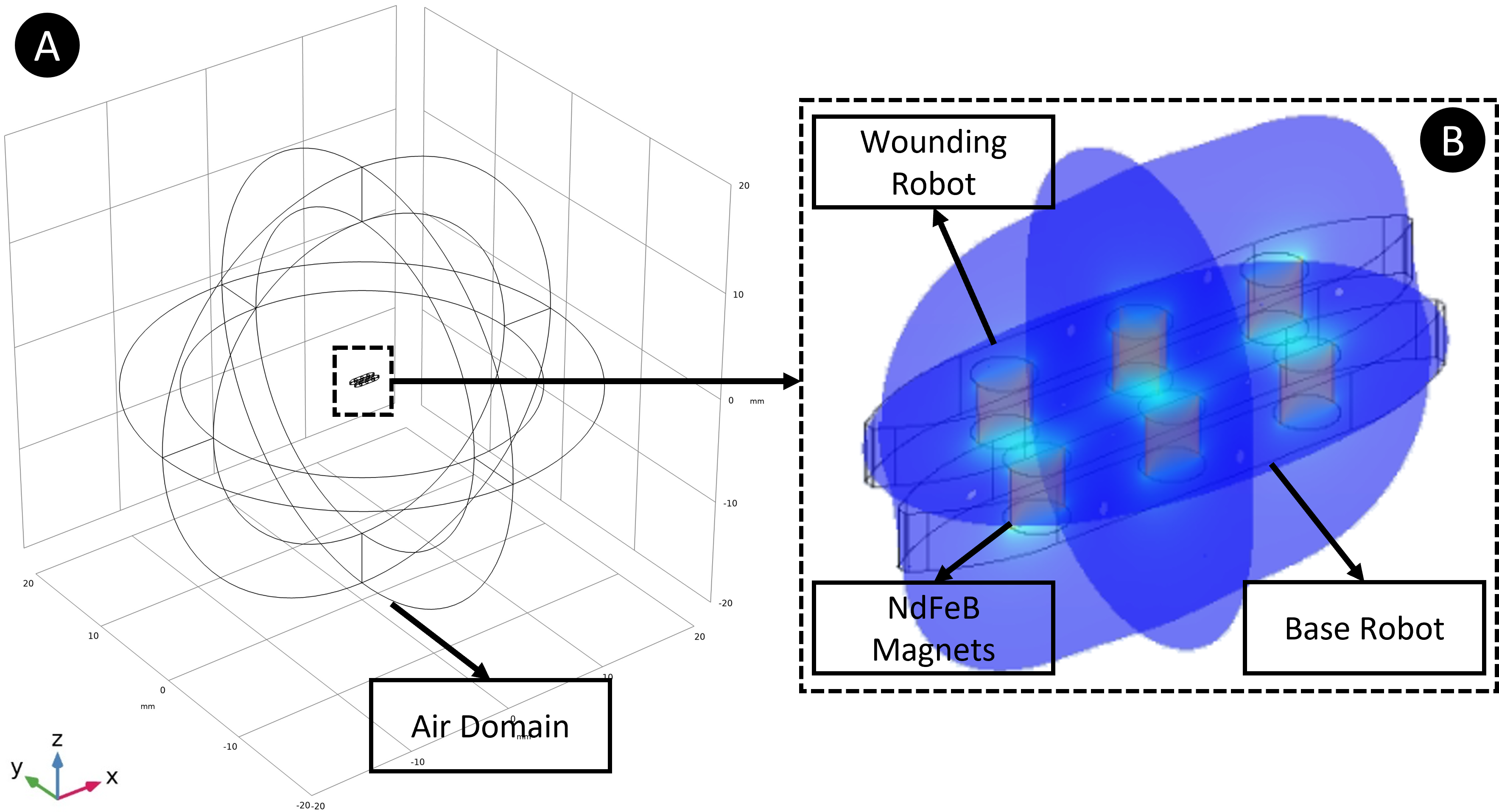}
  \caption{Shows a) overview of the analysis and b) magnetic flux density of 150$\mu$ for the wounding micro robot on the base micro robot by calculating according to the distance between them in air domain.}
  \label{fgr:force_overview}
\end{figure}

Magnetic fields, no currents (mfnc) physics are used with a stationary analysis. The electromagnetic forces between the base robot and the wounding robot are calculated, and the forces in the x and z axes are shown in Fig. \ref{fgr:forcex} and \ref{fgr:forcez}, respectively. It has seen the highest value of 4.64 mN among robots with an offset of 150 $\mu$m in the x-axis. At a distance of 150 $\mu$m between two robots, the electromagnetic force of 5.5 mN occurs in the z direction. Since there is displacement between the robots in x-axis and they stabilize each other in the z-axis, there is no significant relationship in the y-axis. However, since 3 magnets are used in line and symmetrically for the xz plane, the pitch moment cannot be produced. Thanks to this analysis, the offset value between them will be able to be determined depending on the force required to scratch the cell line from the surface in wound healing models.

\begin{figure}[ht]
\centering
  \includegraphics[height=6cm, width=8.8cm]{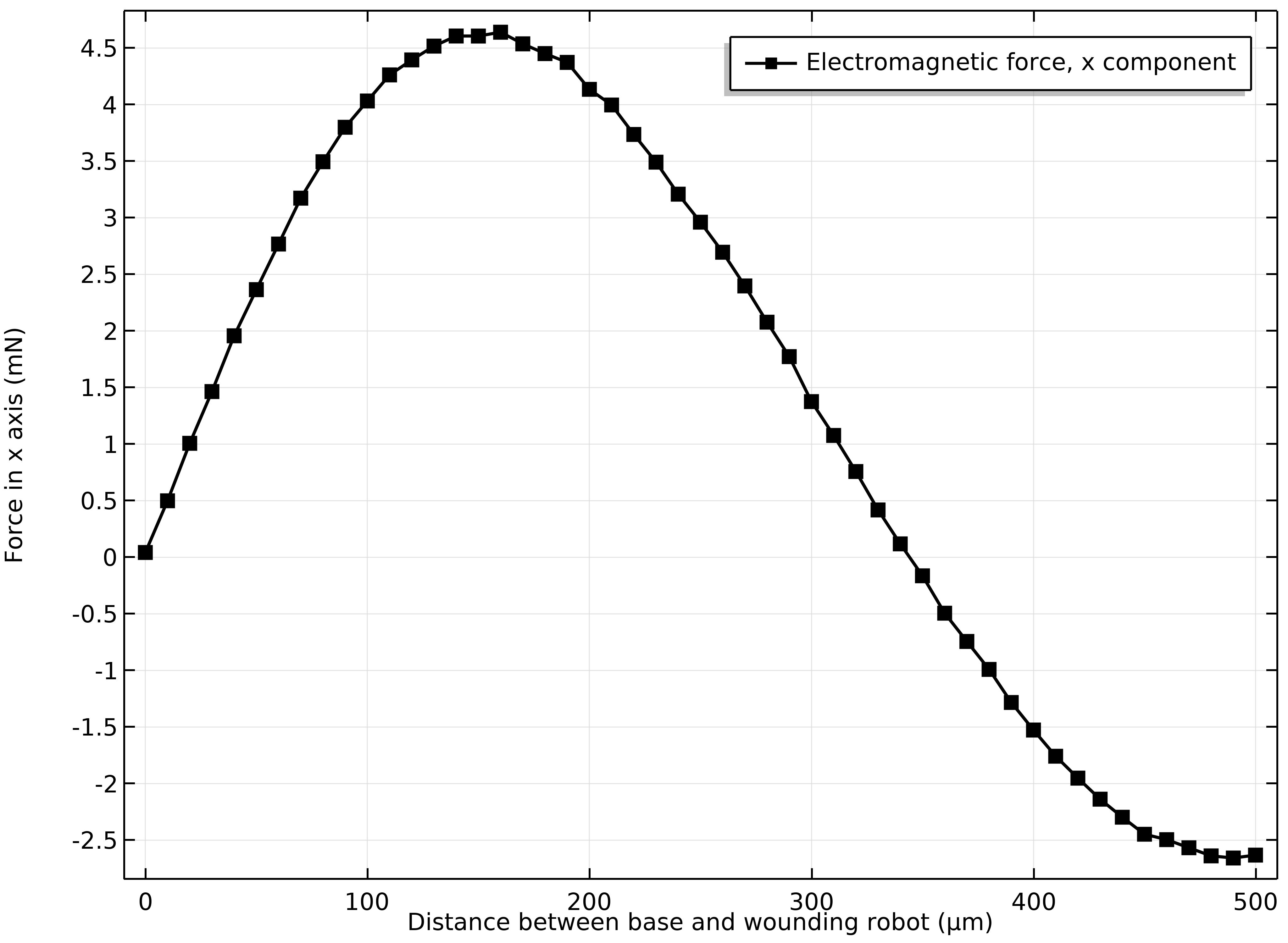}
  \caption{The force curve created by the Wounding micro robot on the base micro robot by calculating according to the distance between them and electromagnetic force on x-axis.}
  \label{fgr:forcex}
\end{figure}


\begin{figure}[ht]
\centering
  \includegraphics[height=6cm, width=8.8cm]{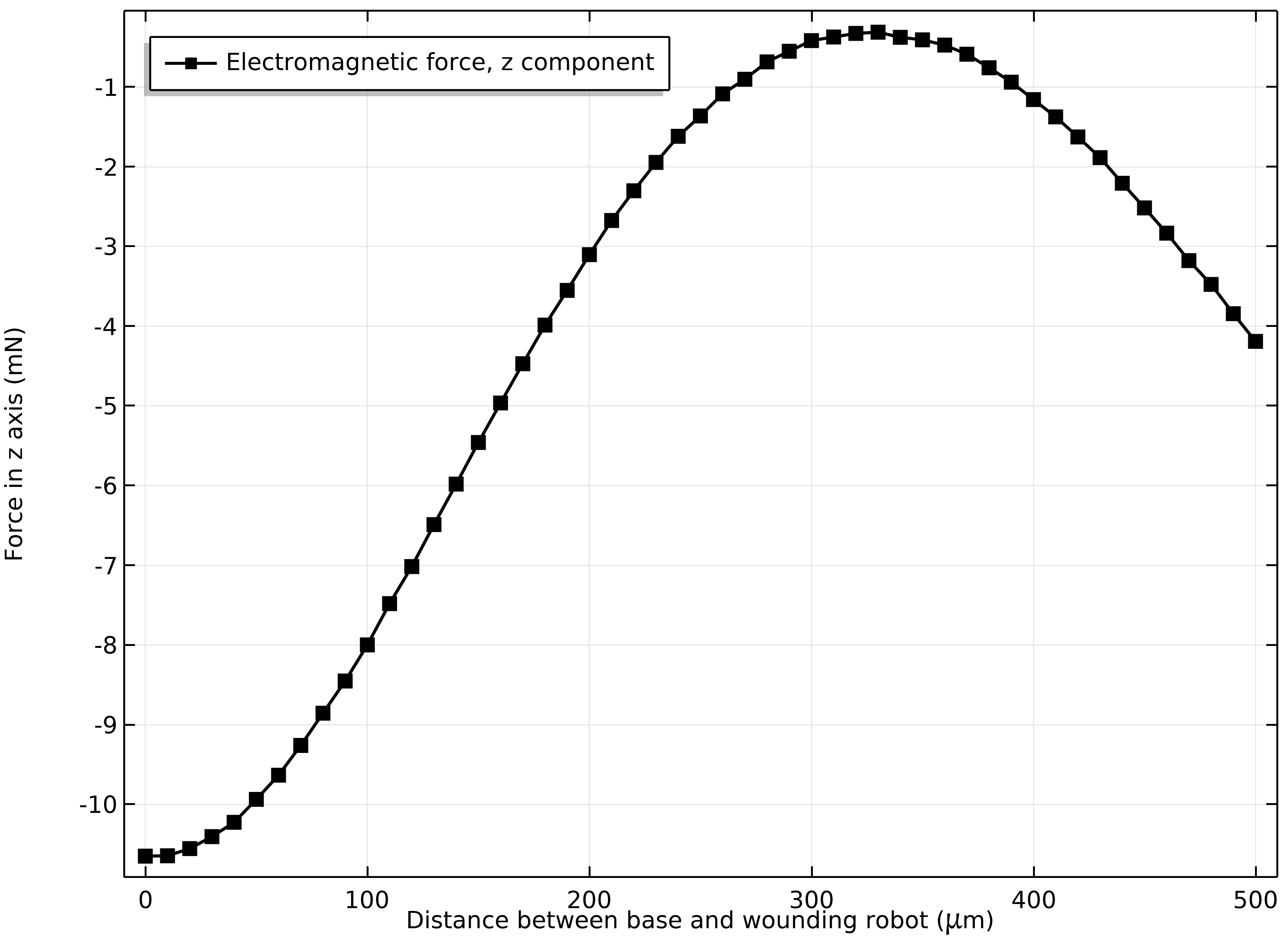}
  \caption{The force curve created by the wounding micro robot on the base micro robot by calculating according to the distance between them and electromagnetic force on z-axis.}
  \label{fgr:forcez}
\end{figure}

\subsection{Friction Force Analysis}
In this study, manipulation in this technique is provided by magnetic interaction in the robot configuration. Unlike, other robot studies that interact with the magnetic field. However, the manipulated robot moves by scraping the surface without being levitated here, as it will perform the surface scraping function. Due to the nonlinear nature of the magnetic field, the electromagnetic force acting on the manipulated robot will not be uniform over time. Therefore, the friction coefficient between the SU-8 body and the glass was chosen as 0.05. Stationary analysis was performed to obtain the initial conditions of the analysis. The geometry of friction analysis was designed that simulates the movement of the robot inside the microchip. The overview of the analysis and the image of the velocity magnitude after the initial conditions are shown in Fig. \ref{fgr:force_overview}. The time-dependent analysis of 1 second was performed with 0.01 second steps. The movement speed of the robot on the surface was chosen as 0.005 m/s. As can be seen from the analysis result, the maximum static friction force is 1.32 nN for the SU-8 robot with the size of 2 mm x 0.25 mm x 0.4 mm.

\begin{figure}[ht]
\centering
  \includegraphics[height=4cm, width=8.8cm]{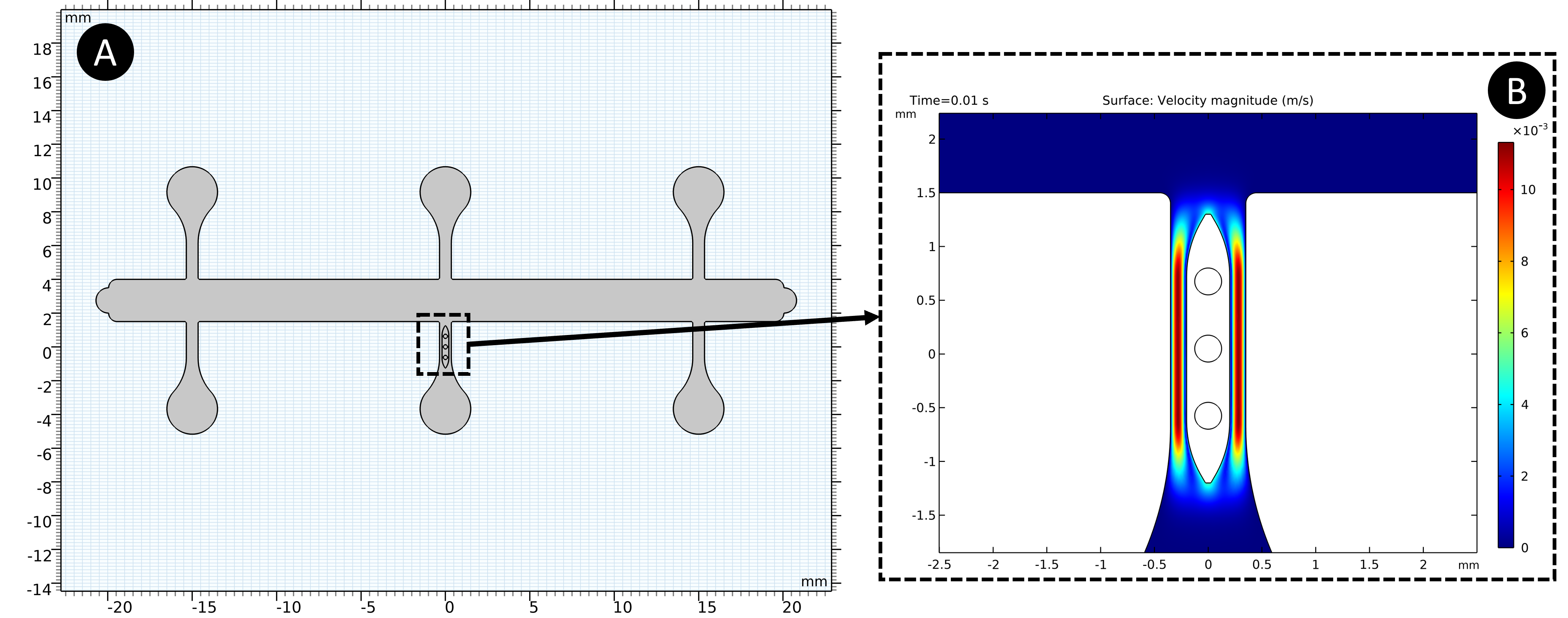}
  \caption {Shows a) overview of friction analysis and b)initial velocity magnitude.}
  \label{fgr:friction_overview}
\end{figure}

\begin{figure}[ht]
\centering
  \includegraphics[height=5cm, width=8.8cm]{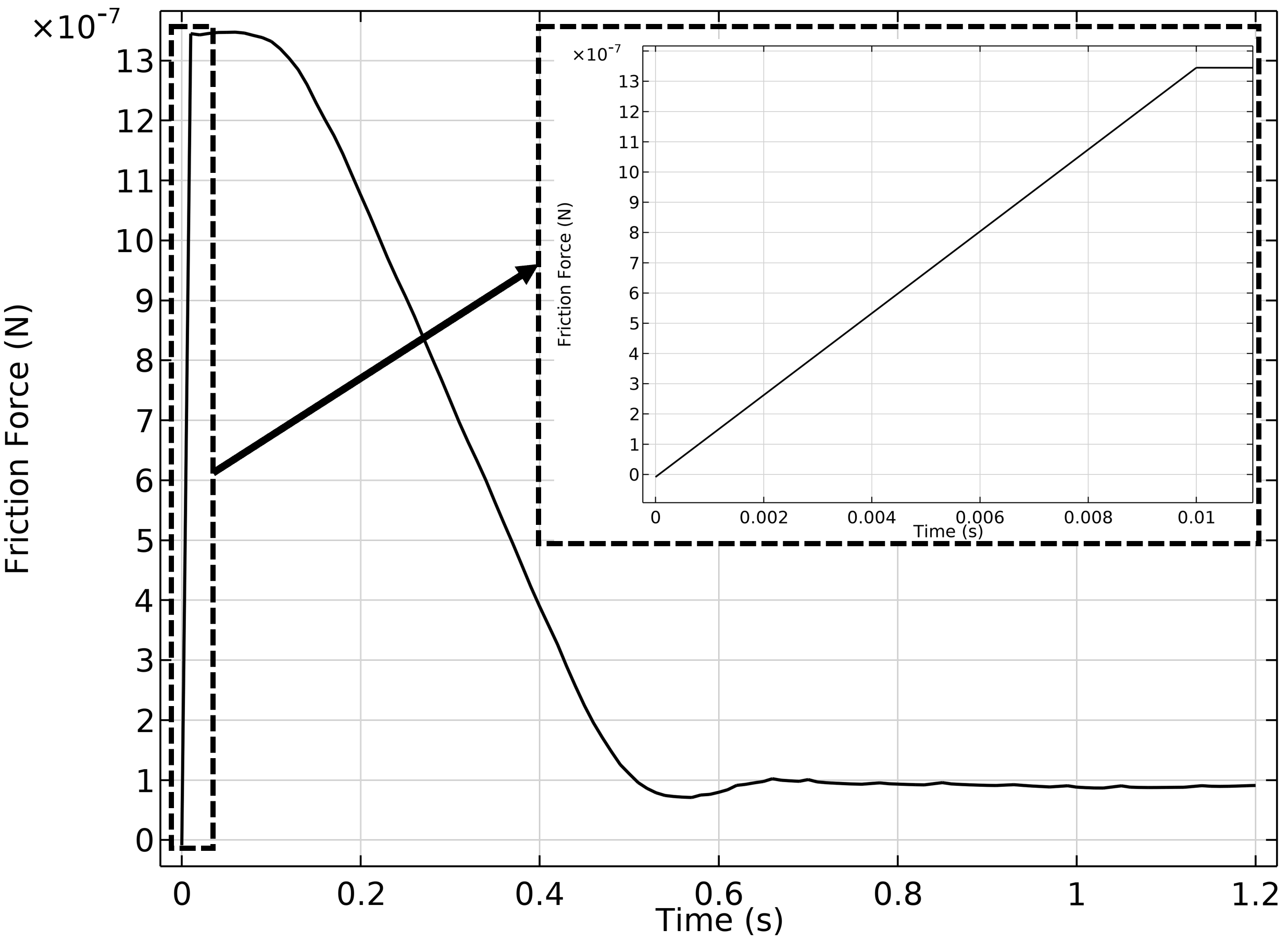}
  \caption{Shows friction force according to time.}
  \label{fgr:friction_force}
\end{figure}

As a result of the analyses, the value that is maximum static friction force in the friction analysis can be obtained at which offset value in the electromagnetic force analysis. The value corresponding to that point in electromagnetic force analysis is the distance between the wounding robot and the base robot. In addition, when the lifting forces of the cells to be used in the wound creating process are added, the distance values that should be between the robots can be calculated from electromagnetic analysis.

\section{Conclusion}
This study demonstrated the required distance between robots for a micro robotic wound opening technique using 3 NdFeB cylindrical shaped permanent magnets with 0.25 mm diameter and 0.25 mm thickness and SU-8 as the body material. The electromagnetic analysis shows electromagnetic forces depending on the distance between the robots. The system can produce a maximum electromagnetic force of 4.64 mN along the x-axis. The graph of the friction force generated by the surface without cells is also shown for a robot with body dimensions of 2 mm x 0.25 mm x 0.4 mm. The friction force and the required force to lift the cells from the surface used in a system allow us to calculate the minimum distance between the robots. In addition to this, the electromagnetic and friction force graphs shows boundaries of the system. By using these graphs, the ideal distance between robots can be obtained experimentally to lift the cells from the surface with the least stress for ideal wound generating process. In future studies, the electromagnetic and friction force graphs for the dynamic cell environment can be studied with the analysis of micro robots in the flow.

\bibliographystyle{ieeetr}
\bibliography{bibliography.bib}
\end{document}